\documentclass[runningheads]{llncs}

\usepackage[T1]{fontenc}
\usepackage{graphicx}
\graphicspath{{../}}
\usepackage{booktabs}
\usepackage{amsmath}
\usepackage{amssymb}
\usepackage{multirow}
\usepackage{xcolor}
\usepackage{hyperref}
\usepackage{float}
\usepackage{subfigure} 
\usepackage{subcaption} 

\newcommand{\seniorauthor}{\textsuperscript{*}}
\begin{document}

\title{MIST: Multimodal Survival Prediction with Genomic-Guided Histology Attention}

\author{
Muhammet Sami Yavuz\inst{1,2}
\and
Sabri Mustafa Kahya\inst{3}
\and
Richard R. Chen\inst{4}
\and
Jana Lipkova\inst{5}\seniorauthor
\and
Benedikt Wiestler\inst{1,2}\seniorauthor
}

\authorrunning{M. S. Yavuz et al.}

\institute{
AI for Image-Guided Diagnosis and Therapy, School of Medicine and Health,\\
Technical University of Munich (TUM), Munich, Germany
\and
Munich Center for Machine Learning (MCML), Munich, Germany
\and
Technical University of Munich (TUM), Munich, Germany
\and
University of California, Irvine, Irvine, CA, USA
\and
Department of Pathology, School of Medicine, and\\
Department of Biomedical Engineering, School of Engineering,\\
University of California, Irvine, Irvine, CA, USA\\[0.5ex]
\email{sami.yavuz@tum.de}\\[0.5ex]
\textsuperscript{*}Shared senior authorship
}

\maketitle

% ─────────────────────────────────────────────────────────────────────────────
% ─────────────────────────────────────────────────────────────────────────────
\begin{abstract}
Multimodal survival models can combine complementary prognostic information from whole-slide images and genomic profiles, but effective fusion remains challenging amid external cohort shift and computational complexity. To address these challenges, we propose \emph{MIST}, \textbf{m}ult\textbf{i}modal \textbf{s}urvival predic\textbf{t}ion with genomic-guided histology attention. MIST represents genomic features as tokens and allows them to query compact foundation-model-derived histology context tokens before survival prediction. This design enriches molecular information with histology context rather than merging separately encoded modalities only at the final stage. Training combines discrete-time survival prediction with genomic feature masking, WSI dropout, and paired WSI–genomics contrastive alignment. Across four external evaluations in colon, renal, lung, and glioblastoma cohorts, MIST improves external C-index over standard fusion baselines in the primary comparisons. These results support genomic-guided histology attention as a compact and effective strategy for multimodal oncology outcome prediction. Our code is available at \url{https://github.com/samiyavuuz/MIST}.

    \keywords{Survival prediction \and Multimodal fusion \and
    Whole slide images \and Genomics \and Cross-Attention}
    \end{abstract}

% ─────────────────────────────────────────────────────────────────────────────
\section{Introduction}

Survival prediction from multimodal cancer data combines complementary prognostic
information from whole-slide histopathology images (WSIs) and molecular profiles that
may not be visible from either modality alone~\cite{chen2021multimodal,chen2022pan}.
Early histology--genomics models were limited by the cost of relating patch-level image
features to high-dimensional molecular measurements, but pathology foundation models
such as TITAN, GIGAPATH, and TANGLE~\cite{ding2024titan,xu2024gigapath,jaume2024transcriptomics} now provide
compact slide-level representations that make WSI/genomics fusion practical across
cohorts. External outcome prediction remains
challenging, however, because models trained on TCGA-like cohorts must generalize to
independent institutions with different patient populations, tissue preparation, slide
availability, and molecular assay designs.

Existing multimodal survival methods demonstrate the value of histology--genomics
fusion but usually define interactions over modality-level embeddings, coarse omic
groups, or predefined pathways. PathOmic~\cite{chen2020pathomic} uses
Kronecker-product fusion of histology graph and genomic features; MCAT~\cite{chen2021multimodal}
introduces co-attention between genomic categories and WSI patch tokens;
PORPOISE~\cite{chen2022pan} scales pan-cancer histology--genomics fusion; and
SurvPath~\cite{jaume2024modeling} models dense pathway--histology interactions. These
designs can limit how individually observed genomic features modulate histology
integration for a given patient. This matters clinically because the prognostic
relevance of morphology may depend on molecular context, and the available genomic
evidence can vary across cohorts.

We present \emph{MIST}, a genomic-guided multimodal survival architecture for joint WSI/genomics outcome prediction. Unlike late-fusion and symmetric co-attention baselines, MIST uses observed genomic feature tokens as queries over compact histology context tokens derived from the TITAN foundation model.
Thus, histology contributes through molecularly conditioned interactions before patient-level survival prediction. Variable-rate genomic masking and WSI dropout regularize training, and paired WSI--genomics InfoNCE alignment encourages matched molecular and histology representations to agree.

We evaluate MIST across four distinct cancer types with varying molecular assays: colon adenocarcinoma, kidney renal clear cell carcinoma, lung squamous cell carcinoma, and glioblastoma. Our contributions are:
\begin{enumerate}
    \item A genomic-guided histology attention module that integrates WSI context into
          the genomic token sequence through feature-level cross-attention.
    \item A multimodal survival training framework combining discrete-time survival
          prediction with paired WSI--genomics InfoNCE alignment.
    \item An external evaluation showing that MIST improves survival prediction over
          standard fusion baselines in various oncology cohorts.
\end{enumerate}
% ─────────────────────────────────────────────────────────────────────────────
\vspace{-5mm}
\section{Method}
\vspace{-3mm}

\begin{figure}[t]
  \centering
  \includegraphics[width=\linewidth]{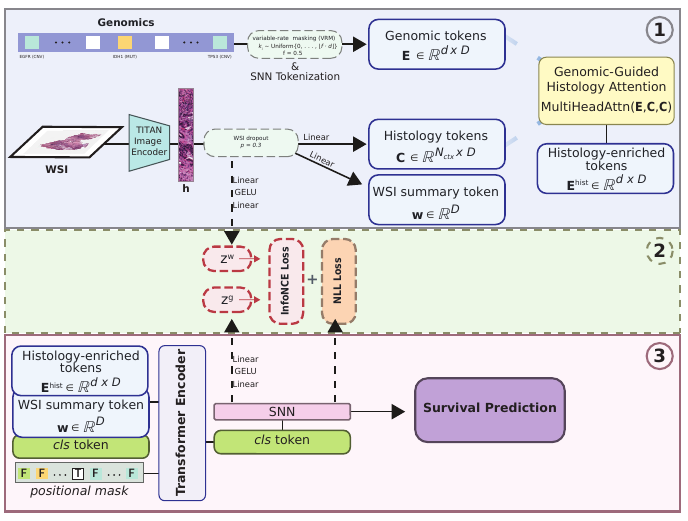}
    \setlength{\abovecaptionskip}{1pt}
    \caption{\scriptsize Overview of MIST. \textbf{(1 - Top)} Genomic features and WSI are separately embedded, and genomic tokens are cross-attention-enriched into $\mathbf{E}^{\mathrm{hist}}$. \textbf{(3 - Bottom)} From a multimodal input sequence, a $\texttt{CLS}$ token is generated through a Transformer encoder, and used for survival prediction. \textbf{(2 - Middle)} Training-only: contrastive alignment of the $\texttt{CLS}$ token and histology embedding. Solid arrows: training/inference forward path. Dashed modules: training-only operations.}
  \label{fig:arch}
\end{figure}

MIST is a multimodal survival model for fusing molecular data with WSI representations. It builds on a missingness-aware genomics-only transformer \cite{shift2026}, with histology-enriched genomic features as a core element (Fig.~\ref{fig:arch}). 
% \subsection{MIST Architecture}
\label{sec:arch}

Let $\mathbf{x}\in\mathbb{R}^{d}$ denote the genomic feature vector for a patient. Each genomic feature is embedded independently as a $D$-dimensional token using a self-normalizing network (SNN)~\cite{klambauer2017self}, implemented as a linear layer followed by SELU activation and AlphaDropout. We set $D=128$ in all experiments, yielding $\mathbf{E}\in\mathbb{R}^{d\times D}$ as the resulting genomic token sequence.

Let $\mathcal{S}=\{s_\ell\}_{\ell=1}^{L}$ denote the set of available WSIs for a patient. To incorporate histology information, we use a WSI encoder $\phi_{\mathrm{WSI}}$ to map each slide $s_\ell$ to a slide-level representation $\mathbf{h}_\ell=\phi_{\mathrm{WSI}}(s_\ell)\in\mathbb{R}^{768}$. In this work, $\phi_{\mathrm{WSI}}$ is instantiated with TITAN~\cite{ding2024titan}. When multiple WSIs are available for a patient, we average their embeddings to obtain a patient-level WSI representation $\mathbf{h}=\frac{1}{L}\sum_{\ell=1}^{L}\mathbf{h}_\ell$. Since TITAN, a slide-level foundation model, is trained to emphasize tumor and tumor-adjacent morphology, each $\mathbf{h}_\ell$ already concentrates tumor-relevant content, and averaging across slides reinforces this shared signal while damping slide-specific noise.

 We use lightweight projection layers to map $\mathbf{h}$ into $N_{\mathrm{ctx}}=4$ compact histology context tokens $\mathbf{C}\in\mathbb{R}^{N_{\mathrm{ctx}}\times D}$ and a global WSI summary token $\mathbf{w}\in\mathbb{R}^{D}$. Both the context-token projection and the summary token $\mathbf{w}$ are obtained using a single linear layer. Using multiple tokens
provides a small learned dictionary of histology components over which
different genomic features can place different attention weights.
We set \(N_{\mathrm{ctx}}=4\) as a compact bottleneck that permits
feature-specific interactions while keeping the number of fusion parameters
and the computational cost small. This choice also matches the attention head count inherited from the SHIFT genomics-only backbone~\cite{shift2026}, keeping the histology context consistent in scale with the rest of the architecture rather than introducing an independently tuned hyperparameter.

MIST uses genome-guided cross-attention to enrich the genomic token sequence prior to the transformer encoder. Specifically, genomic tokens serve as queries, while the histology context tokens serve as keys and values:
\begin{equation}
\mathbf{E}^{\mathrm{hist}} =
\mathrm{MultiHeadAttn}(\mathbf{E}, \mathbf{C}, \mathbf{C}).
\end{equation}
This yields histology-enriched genomic tokens $\mathbf{E}^{\mathrm{hist}}\in\mathbb{R}^{d\times D}$, allowing each genomic feature token to attend to and be enriched by compact slide-level context.

For patient $i$, the full transformer input is formed as
$\mathbf{Z}_i=[\texttt{CLS}, \mathbf{w}_i, \mathbf{E}^{\mathrm{hist}}_i]$.
The \texttt{[CLS]} token aggregates the final patient representation, $\mathbf{w}_i$
provides a global WSI summary, and $\mathbf{E}^{\mathrm{hist}}_i$ carries
feature-level genomic evidence enriched by histology context. This sequence is processed by a two-layer transformer encoder with four attention heads. Let
$\mathbf{g}_i=T(\mathbf{Z}_i)_0$ denote the final \texttt{[CLS]} output, which is
passed through a SELU-activated linear projection with dropout ($D{\to}256$) to
obtain $\tilde{\mathbf{g}}_i$. The survival head produces
$\hat{\mathbf{y}}_i=\mathbf{W}_o\tilde{\mathbf{g}}_i+\mathbf{b}_o$, which is converted into discrete survival hazard probabilities over $K=4$ time
intervals using the discrete-time negative log-likelihood (NLL, see Sec.~\ref{sec:nll})
loss~\cite{zadeh2020survival}.

\vspace{-4mm}
\subsection{Training-Time Regularization}

\vspace{-2mm}
MIST uses two training-time regularization mechanisms to improve robustness to incomplete inputs. \textbf{First}, we apply variable-rate masking (VRM) to regularize genomic encoding and reduce overfitting to dominant genomic features. Natural genomic missingness is represented by the original availability mask, while VRM introduces an additional stochastic training-time mask over observed features. For each patient and epoch, exact-$k$ VRM samples $k_i \sim \mathrm{Uniform}\{0,\ldots,\lfloor f\cdot d\rfloor\}$ observed genomic feature positions and masks them in addition to naturally missing features (we set $f=0.5$).

\textbf{Second}, we apply WSI dropout to regularize histology enrichment and reduce overfitting to dominant WSIs. With probability $p_{\text{wsi}}=0.3$, input WSI representations, i.e., TITAN embeddings, are omitted during training. Dropped WSIs bypass histology enrichment, and their corresponding WSI summary tokens are masked in the transformer input. Thus, VRM targets feature-level genomic missingness, while WSI dropout acts as a slide-level regularizer. Missing inputs are handled through binary availability masks throughout the architecture. Naturally absent genomic features and VRM-selected features are excluded from transformer self-attention. Similarly, when no WSI is available for a patient, or when a WSI representation is omitted by WSI dropout, genome-guided histology attention is skipped, the WSI summary token is masked in the transformer input, and the model proceeds using the remaining available tokens.

\subsection{Survival Objective and Alignment Loss}
\label{sec:nll}
We model survival in discrete time, following the same formulation as in \cite{shift2026}. For each training fold, continuous survival times are
partitioned into $K=4$ intervals using quartiles of uncensored event times. For each
interval $j\in\{0,\ldots,K-1\}$, the network outputs a logit $\hat{y}_j$, which is
converted into a hazard probability $\hat{h}_j=\sigma(\hat{y}_j)$. The predicted survival
probability through interval $j$ is \vspace{-3mm}
\begin{equation}
    S_j = \prod_{m=0}^{j}(1-\hat{h}_m), \qquad S_{-1}=1.
\end{equation}

%For a patient assigned to interval $Y\in\{0,\ldots,K-1\}$, let $c\in\{0,1\}$ denote the censoring indicator, where $c=1$ indicates right-censoring and $c=0$ indicates an observed event.
The discrete-time survival loss is formulated using one censored term and two event terms to capture the negative log-likelihood of the observed data. For C-index computation, we use the scalar risk score $r=-\sum_{j=0}^{K-1}S_j$.

MIST also uses a symmetric InfoNCE alignment loss~\cite{oord2018representation} to align the survival representation with the paired WSI representation. Let $\mathbf{g}_i$ denote the final survival \texttt{[CLS]} representation for patient $i$, and let $\mathbf{h}_i$ denote the corresponding TITAN WSI embedding. Two MLP projection heads, $q_g$ and $q_w$, each implemented as Linear--GELU--Linear, map these vectors into a shared $D$-dimensional contrastive space and produce normalized embeddings $\mathbf{z}^{g}_i = q_g(\mathbf{g}_i)/\lVert q_g(\mathbf{g}_i)\rVert_2$ and $\mathbf{z}^{w}_i = q_w(\mathbf{h}_i)/\lVert q_w(\mathbf{h}_i)\rVert_2$. For a minibatch subset of $N$ patients with paired genomic and WSI inputs, we compute the temperature-scaled similarity matrix $A_{ij}=((\mathbf{z}^{g}_i)^\top \mathbf{z}^{w}_j)/\tau$, where $i,j\in{1,\ldots,N}$, and $\tau$ is the InfoNCE temperature. The diagonal entries correspond to matched WSI-genomics pairs from the same patient,
while off-diagonal entries are in-batch negatives: \begin{equation}
    \mathcal{L}_{\mathrm{InfoNCE}}
    = \frac{1}{2}
    \left[
    \frac{1}{N}\sum_{i=1}^{N}
    -\log\frac{\exp(A_{ii})}{\sum_{j=1}^{N}\exp(A_{ij})}
    +
    \frac{1}{N}\sum_{i=1}^{N}
    -\log\frac{\exp(A_{ii})}{\sum_{j=1}^{N}\exp(A_{ji})}
    \right].
\end{equation}
This loss is computed only when at least two minibatch samples have both genomic and
WSI inputs. The final training objective is
\begin{equation}
    \mathcal{L} =
    \mathcal{L}_{\mathrm{NLL}}
    +
    \mathcal{L}_{\mathrm{InfoNCE}},
    \qquad
    \tau=0.07.
\end{equation}

\vspace{-3mm}

% ─────────────────────────────────────────────────────────────────────────────
\vspace{-4mm}
\section{Experiments}
\vspace{-2mm}
\subsection{Datasets}

We evaluate MIST across four histomorphologically distinct cancer types with various molecular assays: colon adenocarcinoma (COAD), kidney renal clear cell carcinoma (KIRC), lung squamous cell carcinoma (LUSC), and glioblastoma (GBM). TCGA cohorts are used for model development, while CPTAC~\cite{edwards2015cptac} and independent institutional cases serve as held-out external tests. TITAN slide-level features are precomputed for all WSIs. We preserve cancer-specific genomic feature spaces from LinkedOmics multi-omics\cite{vasaikar2018linkedomics} rather than imposing a pan-cancer intersection. The available molecular data span mutations and copy number variation (CNV) profiles. Across cohorts, TCGA and external cases are aligned by gene symbol and case identifier, and unmeasured values remain NaNs rather than being imputed, so all models learn from observed genomic evidence and patient-level TITAN WSI embeddings. Full dataset statistics, modality availability, and genomic feature spaces are summarized in Table~\ref{tab:datasets}.

\textit{Genomic feature curation:}
For COAD, we use 143 recurrently mutated genes, selected as those mutated in at least 10\% of COAD-TCGA samples, together with the 500 genes with the highest variance in GISTIC2 thresholded CNV calls across COAD-TCGA samples. All features are restricted to genes shared between the TCGA and CPTAC cohorts. Among COAD-TCGA patients, 5 are mutation-only, and 81 are CNV-only, whereas all COAD-CPTAC patients have both. Notably, COAD-CPTAC has a high censoring rate of 93.1\%, corresponding to approximately seven observed events; results on this cohort should therefore be interpreted with caution. KIRC has 920 matched KIRC-TCGA and KIRC-CPTAC gene-level copy-number log$_2$ ratio features. CNVs are selected using a 10\% CNV-frequency threshold, restricted to genes shared between TCGA and CPTAC.
 GBM-TCGA and GBM-GERMAN include IDH1 mutation status and 35 clinically relevant CNV features, including 1p/19q codeletion. These genes were selected based on genes predefined in the conumee R package \cite{conumee} and enriched with additional CNV alterations described in \cite{touat2017glioblastoma}. LUSC-TCGA has 197 recurrently mutated genes, whereas LUSC-US contains only 22, with the remaining genes missing. For all cancers, we apply z-score normalization to all genomic features, using the mean and standard deviation computed from the training data.

\subsection{Baselines and Evaluation Protocol}

All fusion baselines use the same missingness-aware genomic backbone, TITAN embeddings,
discrete survival loss, VRM setting, and WSI dropout probability as MIST; they differ
only in the fusion operator. Missing genomic features remain NaNs and are excluded by
the shared backbone mask, while missing WSI is represented by a zero TITAN vector.
InfoNCE alignment is used only by MIST.

\textbf{Concat}~\cite{chen2022pan} projects the TITAN embedding to the CLS dimension
with a linear-ReLU layer, concatenates it with the genomic \texttt{[CLS]} token, and
uses a two-layer MLP survival head ($512{\to}256{\to}K$, dropout 0.25).
\textbf{Bilinear} adapts PORPOISE gated outer-product fusion~\cite{chen2022pan}: each
modality is sigmoid-gated using both embeddings, and the outer product of the gated
vectors is compressed to 256 dimensions before survival prediction. \textbf{Co-Attn}
implements MCAT-style bidirectional co-attention~\cite{chen2021multimodal}, with the
genomic \texttt{[CLS]} token attending to the projected TITAN embedding and the TITAN
embedding attending to the genomic \texttt{[CLS]} token; the two enriched outputs are
concatenated before a two-layer MLP survival head. \textbf{Mask-aware variants} distinguish true modality absence from zero-valued
embeddings. Mask-aware concat appends
binary genomics- and WSI-presence indicators to the fused representation;
mask-aware bilinear uses bilinear fusion only when both modalities are present and
unimodal/default heads otherwise; mask-aware co-attention appends the same indicators
and skips co-attention when either modality is absent.

In addition to the fusion baselines, we include unimodal baselines. \textbf{Genomics-only} and \textbf{WSI-only} analyze only one modality at a time with no fusion operator to isolate the advantage of modality fusion. To this end, we chose configurations similar to fusion baselines: \textbf{Genomics-only} passes the genomic \texttt{[CLS]} token through a two-layer survival head of the same shape as \textbf{Concat}'s genomic branch ($256{\to}256{\to}K$, dropout 0.25). It uses the same missingness-aware genomic backbone and VRM feature masking. \textbf{WSI-only} projects the TITAN embedding through the same Linear--ReLU layer and survival-head shape as \textbf{Concat}'s WSI branch.

All models are trained on TCGA with stratified 5-fold cross-validation and evaluated on external cohorts. We use Adam \cite{kingma2014adam} with learning rate $5\times10^{-4}$, weight
decay $10^{-5}$, batch size 16, up to 200 epochs, and early stopping
based on validation loss (survival NLL plus $\ell_1$ weight regularization).

\vspace{-5mm}
\begin{table}[H]
\centering

\caption{\scriptsize
  TCGA development and external test cohorts for multimodal survival
  evaluation. \textbf{Slides} counts WSIs, allowing multiple slides per patient;
  \textbf{Paired}, \textbf{WSI-only}, and \textbf{Gen-only} denote modality
  availability; Cens.\,\% is the censored fraction. The final row is the paired
  GBM control restricted to paired multimodal cases.}
\label{tab:datasets}
\resizebox{\textwidth}{!}{%
\begin{tabular}{@{}llr rrrrr r | l rrrrr r@{}}
\toprule
\multirow{2}{*}{Cancer} & \multirow{2}{*}{Genomic Type} & \multirow{2}{*}{\#Feat} &\multicolumn{6}{c|}{Training (TCGA, train\,+\,val)} &\multicolumn{7}{c}{External Test Cohort} \\
\cmidrule(lr){4-9}\cmidrule(l){10-16}
& & & Patients & Slides & Paired & WSI-only & Gen-only & Cens.\,\% & Cohort & Patients & Slides & Paired & WSI-only & Gen-only & Cens.\,\% \\
\midrule
COAD & 143 Mut + 500 CNV & 643 & 423 & 429 & 423 & 0 & 0 & 78.0\% & CPTAC & 101 & 214 & 101 & 0 & 0 & 93.1\% \\
\addlinespace[2pt]
GBM & 1 Mut + 35 CNV & 36 & 532 & 858 & 276 & 112 & 144 & 20.1\% & German & 132 & 132 & 132 & 0 & 0 & 68.9\% \\
\addlinespace[2pt]
KIRC & 920 CNV & 920 & 488 & 493 & 488 & 0 & 0 & 65.8\% & CPTAC & 94 & 319 & 94 & 0 & 0 & 77.7\% \\
\addlinespace[2pt]
LUSC & 197 Mut & 197 & 465 & 499 & 465 & 0 & 0 & 57.6\% & US & 102 & 102 & 102 & 0 & 0 & 46.1\% \\
\midrule
\multicolumn{16}{l}{\textit{Matched-cohort control (WSI+genomics required for all patients)}} \\
\addlinespace[2pt]
GBM (paired) & 1 Mut + 35 CNV & 36 & 276 & 643 & 276 & 0 & 0 & 14.1\% & German & 132 & 132 & 132 & 0 & 0 & 68.9\% \\
\bottomrule
\end{tabular}%
}

\end{table}

% ─────────────────────────────────────────────────────────────────────────────

\section{Results}

\subsection{MIST Outperforms Established Baselines Across Cancer Types and Molecular Arrays}

Table~\ref{tab:main} compares MIST with standard fusion operators under identical training and feature settings. In the four primary external evaluations, MIST achieves the highest C-index and tAUC across all cohorts. The strongest gains appear in cohorts where external generalization and molecular heterogeneity are most challenging. On COAD-CPTAC, MIST improves C-index by 9.3\% relative to the best non-MIST comparator and improves tAUC by 9.6\%. In LUSC-US, where the external mutation panel is compositionally sparser than the TCGA source panel, MIST improves C-index by 7.1\% over the best baseline and tAUC by 8.6\%. These results suggest that conditioning histology integration on observed genomic features improves external risk ranking beyond late fusion, bilinear fusion, and symmetric co-attention. Importantly, the unimodal baselines highlight the value of modality fusion in MIST: across all four primary external evaluations, MIST outperforms both the Genomics-only and WSI-only models. This advantage is particularly pronounced over Genomics-only, which hovers near chance level (C-index and tAUC $\approx 0.50$) in every cohort, suggesting that fusing histology context into the genomic tokens is what allows the model to learn a transferable risk representation.

The paired GBM-German control in Table~\ref{tab:main} corresponds to the final row of Table~\ref{tab:datasets}. For this control, all preprocessing steps, including survival-bin construction, are recomputed using only patients with both WSI and genomic measurements; MIST and all fusion baselines are then retrained with the primary optimization settings
Crucially, restricting the cohort to fully paired cases not only reduces the available training data from 532 to 276 patients but also eliminates the modality-level heterogeneity that MIST's genomic-guided histology attention, VRM, and WSI dropout were specifically designed to leverage. Rather than indicating a fundamental limitation, MIST’s performance relative to symmetric co-attention on this subset underscores its specialized design paradigm: MIST trades marginal performance gains on small, artificially complete datasets for robust, superior generalization on inherently incomplete, real-world cohorts.
\vspace{-2mm}
 \begin{table}[H]
    \centering
    \setlength{\abovecaptionskip}{4pt}
\caption{\scriptsize Held-out external test performance for MIST and fusion/unimodal baselines. Each cell
reports C-index | tAUC as $\text{mean}_{\pm \text{std}}$ across five folds. Both metrics are
higher-is-better. ``-M'' denotes the mask-aware version of
the corresponding fusion baseline.}
    \label{tab:main}
    \setlength{\tabcolsep}{1pt}
    \resizebox{\textwidth}{!}{%
    \begin{tabular}{lcccc|c}
    \toprule
    Method & COAD-CPTAC & GBM-German & KIRC-CPTAC & LUSC-US & GBM-German paired \\
    \midrule
    Concat & $0.581_{\pm0.06} | 0.653_{\pm0.06}$ & $0.473_{\pm0.04} | 0.451_{\pm0.05}$ & $0.565_{\pm0.05} | 0.559_{\pm0.06}$ & $0.529_{\pm0.04} |0.517_{\pm0.06}$ & $0.489_{\pm0.04} |  0.499_{\pm0.07}$ \\
    Bilinear & $0.538_{\pm0.08} | 0.608_{\pm0.11}$ & $0.483_{\pm0.03} | 0.470_{\pm0.04}$ & $0.577_{\pm0.04} | 0.578_{\pm0.05}$ & $0.510_{\pm0.02} |0.513_{\pm0.03}$ & $0.533_{\pm0.05} |  0.550_{\pm0.08}$ \\
    Co-Attn & $0.541_{\pm0.12} | 0.616_{\pm0.15}$ & $0.483_{\pm0.05} |  0.485_{\pm0.06}$ & $0.562_{\pm0.04} |0.568_{\pm0.04}$ & $0.524_{\pm0.03} |  0.518_{\pm0.06}$ & $\mathbf{0.555}_{\pm0.04} |  \mathbf{0.580}_{\pm0.05}$ \\
    Concat-M & $0.558_{\pm0.10} |0.617_{\pm0.16}$ & $0.473_{\pm0.03} |  0.455_{\pm0.04}$ & $0.575_{\pm0.05} | 0.559_{\pm0.07}$ & $0.522_{\pm0.05} |  0.504_{\pm0.07}$ & $0.512_{\pm0.03} |  0.535_{\pm0.05}$ \\
    Bilinear-M & $0.479_{\pm0.08} | 0.536_{\pm0.06}$ & $0.568_{\pm0.05} |0.589_{\pm0.08}$ & $0.602_{\pm0.04} |  0.598_{\pm0.06}$ & $0.533_{\pm0.03} |  0.558_{\pm0.06}$ & $0.514_{\pm0.05} |0.537_{\pm0.07}$ \\
    Co-Attn-M & $0.612_{\pm0.07} | 0.711_{\pm0.09}$ & $0.492_{\pm0.01} | 0.508_{\pm0.04}$ & $0.571_{\pm0.04} |  0.564_{\pm0.06}$ & $0.497_{\pm0.05} |0.477_{\pm0.08}$ & $0.521_{\pm0.04} |  0.549_{\pm0.07}$ \\
        Genomics-only & $0.510_{\pm0.03} | 0.490_{\pm0.02}$ & $0.502_{\pm0.01} | 0.503_{\pm0.02}$ & $0.500_{\pm0.00} | 0.500_{\pm0.00}$ & $0.489_{\pm0.02} | 0.478_{\pm0.05}$ & $0.503_{\pm0.02} | 0.523_{\pm0.03}$ \\
    WSI-only & $0.546_{\pm0.04} | 0.685_{\pm0.03}$ & $0.445_{\pm0.03} | 0.417_{\pm0.04}$ & $0.627_{\pm0.03} | 0.630_{\pm0.03}$ & $0.545_{\pm0.01} | 0.549_{\pm0.03}$ & $0.541_{\pm0.04} | 0.558_{\pm0.06}$ \\
    \textbf{MIST} & $\mathbf{0.669}_{\pm0.07} |  \mathbf{0.779}_{\pm0.09}$ & $\mathbf{0.584}_{\pm0.05} |  \mathbf{0.607}_{\pm0.06}$ & $\mathbf{0.646}_{\pm0.02} |\mathbf{0.651}_{\pm0.02}$ & $\mathbf{0.571}_{\pm0.03} |  \mathbf{0.606}_{\pm0.05}$ & $0.491_{\pm0.08} |  0.493_{\pm0.11}$ \\
    \bottomrule
    \end{tabular}}
    \end{table}

\vspace{-10mm}
\subsection{MIST Produces Prognostically Relevant Risk Separation}

\begin{figure}[!htbp]
\centering
\subfigure[]{
  \includegraphics[width=0.38\textwidth]{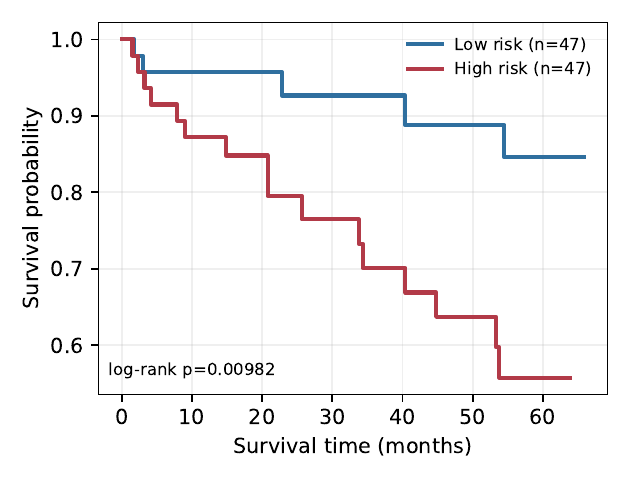}
}
\hspace{-0.5em}
\subfigure[]{
  \includegraphics[width=0.38\textwidth]{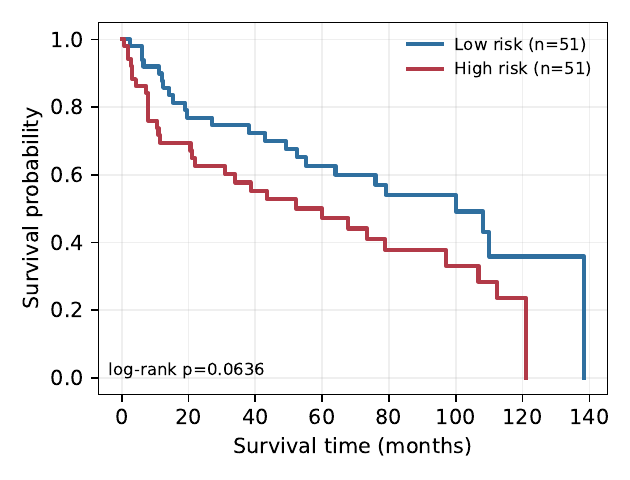}
}
\setlength{\abovecaptionskip}{2pt}
\caption{\scriptsize Kaplan--Meier curves from MIST risk stratification. Patients are dichotomized
at the median averaged risk across the five-fold-trained models. \textbf{(a)} KIRC-CPTAC shows significant separation ($p=0.0098$), while
\textbf{(b)} LUSC-US shows trend-level separation ($p=0.0636$) in the cohort with the strongest structural genomic feature mismatch.}
\label{fig:km_examples}
\end{figure}

We next assessed whether ensemble-averaged MIST risk scores separate external patients into prognostic groups. Table~\ref{tab:mist_risk_stratification} reports patient-level C-index, bootstrap confidence intervals, and Kaplan--Meier log-rank tests. 
The clearest risk-stratification signal is observed in KIRC-CPTAC, where median-risk groups separate significantly by log-rank test (Fig.~\ref{fig:km_examples}, left). LUSC-US shows a trend toward separation (Fig.~\ref{fig:km_examples}, right), consistent with the main external ranking result in the most structurally incomplete molecular setting. Several cohorts have moderate patient-level C-indices but non-significant log-rank tests, particularly where event counts are small, underscoring that risk-separation tests are more sensitive to cohort size and censoring than fold-wise ranking metrics.
These analyses support the prognostic relevance of MIST scores but are complementary to the fold-wise C-index comparisons rather than definitive clinical stratification.
\vspace{-10mm}
\begin{table}[!htbp]
\centering
\caption{\scriptsize MIST risk stratification on external cohorts. Fold-averaged
risk scores yield a patient-level C-index (bootstrap 95\% CI) and a median-split
log-rank test, distinct from the fold-wise comparison in Table~\ref{tab:main}.}
\label{tab:mist_risk_stratification}
\scriptsize
\setlength{\tabcolsep}{3pt}
\begin{tabular}{llccc}
\toprule
Cancer & Cohort & $N$ & C-index (95\% CI) & Log-rank $p$ \\
\midrule
COAD & CPTAC & 101 & $0.699$ ($0.503$--$0.895$) & $0.4478$ \\
GBM & German & 132 & $0.586$ ($0.488$--$0.680$) & $0.1594$ \\
KIRC & CPTAC & 94 & $0.651$ ($0.524$--$0.784$) & $\mathbf{0.0098}$ \\
LUSC & US & 102 & $0.595$ ($0.513$--$0.671$) & $0.0636$ \\
\bottomrule
\end{tabular}
\end{table}

\vspace{-10mm}

\newpage
\subsection{Ablations Support Genomic-Guided Histology Attention and Alignment}

Finally, we ablated key components of MIST on KIRC-CPTAC and LUSC-US, two cohorts with
clear external signal in the main evaluation (see Table \ref{tab:ablations}). Each ablation removes one component while
keeping the same training and evaluation protocol: genomic-guided histology attention,
the InfoNCE alignment term, variable-rate genomic masking (VRM), or WSI dropout.

The ablations identify the genomic-guided histology path and InfoNCE alignment as the
most important contributors. Removing histology attention decreases C-index by 15.3\%
on KIRC-CPTAC and 7.2\% on LUSC-US, while removing InfoNCE decreases C-index by
10.5\% and 13.5\%, respectively. WSI dropout has a smaller effect: it modestly improves
KIRC-CPTAC robustness, but has little impact on LUSC-US. VRM is similarly
cohort-dependent, with minimal effect on KIRC-CPTAC and a slight decrease relative to
the no-VRM variant on LUSC-US. Overall, these results support MIST's central design:
observed genomic features should guide histology integration, and paired WSI--genomics
alignment helps stabilize the learned multimodal representation.

\vspace{-4mm}
\begin{table}[!htbp]
\centering
\caption{\scriptsize Component ablations on selected external cohorts, reported as C-index with
standard deviation as a subscript across five folds. Higher is better.}
\label{tab:ablations}
\setlength{\tabcolsep}{3.0pt}
\scriptsize
\resizebox{0.88\textwidth}{!}{%
\begin{tabular}{lccccc}
\toprule
Cohort & \textbf{Full MIST} & No Hist. Attn & No InfoNCE & No VRM & No WSI Dropout \\
\midrule
KIRC-CPTAC & $0.646_{\pm0.02}$ & $0.546_{\pm0.04}$ & $0.578_{\pm0.04}$ & $0.642_{\pm0.01}$ & $0.610_{\pm0.07}$ \\
LUSC-US & $0.571_{\pm0.03}$ & $0.530_{\pm0.06}$ & $0.494_{\pm0.02}$ & $0.577_{\pm0.05}$ & $0.569_{\pm0.05}$ \\
\bottomrule
\end{tabular}}
\end{table}

% ─────────────────────────────────────────────────────────────────────────────
\vspace{-6mm}
\section{Discussion and Conclusion}
\vspace{-2mm}

In summary, MIST provides a compact strategy for WSI/genomics survival prediction in which observed genomic evidence guides the integration of histology context. Across multiple external oncology cohorts, this genomic-guided attention mechanism improves C-index and tAUC over standard fusion baselines. It also demonstrates clear gains in shifted or in the most systematically incomplete settings, such as COAD-CPTAC, GBM-German, KIRC-CPTAC, and LUSC-US.
The ablation results suggest that MIST’s performance arises from complementary mechanisms that promote biologically meaningful fusion and robustness under cohort shift. First, the histology-enriched genomic tokens allow each observed genomic feature to retrieve the morphological context most relevant to it, rather than compressing both modalities independently and combining them only at the patient level. Second, the InfoNCE objective explicitly encourages the learned survival representation to remain consistent with the matched WSI while distinguishing it from other patients’ histology. We speculate that this patient-level alignment discourages modality-specific shortcuts and stabilizes the shared representation across assay and institutional differences, consistent with the pronounced performance loss when InfoNCE was removed (Table \ref{tab:ablations}). Finally, VRM exposes the model to varying subsets of genomic observations during training \cite{shift2026}. This discourages reliance on a small number of dominant features and encourages the model to distribute evidence across the available genomic and histological inputs, which may be particularly valuable when external cohorts use incomplete or compositionally different molecular panels. Collectively, these findings suggest that MIST benefits not from fusion alone, but from combining feature-specific histology conditioning, cross-modal representation alignment, and robustness to variable genomic availability.
%The matched GBM-German control and MIST-only multi-seed risk review also show that some gains are cohort- and seed-sensitive rather than universal.
While these performance gains highlight MIST's strengths, the current study is limited by modest external cohort sizes and the exclusion of standard clinical covariates.
Future research will address these constraints by integrating explicit patient metadata into the model and incorporating imputation-based baselines. Overall, MIST provides a compact and effective approach for using genomic evidence to guide histology integration in multimodal cancer survival prediction.

% ─────────────────────────────────────────────────────────────────────────────

\bibliographystyle{splncs04}
\bibliography{mybibliography}

\end{document}